\documentclass[reqno]{amsart}

\usepackage{hyperref}
\usepackage[dvipsnames]{xcolor}
\definecolor{burntorange}{HTML}{BF5700}
\definecolor{UTblue}{HTML}{00A9B7}
\definecolor{bluebonnet}{HTML}{005F86}
\hypersetup{
    pdftitle={Architecture independent generalization bounds for overparametrized deep ReLU networks},
    linktoc=all,     
    colorlinks=true, 
    linkcolor=bluebonnet,
    citecolor=burntorange,
    urlcolor=bluebonnet,
}

\usepackage{mathtools}

\def\R{{\mathbb R}}
\def\Z{{\mathbb Z}}

\def\cC{{\mathcal C}}
\def\cD{{\mathcal D}}
\def\cE{{\mathcal E}}

\def\cI{{\mathcal I}}

\def\cL{{\mathcal L}}

\def\cS{{\mathcal S}}

\def\1{{\bf 1}}

\def\Z{\theta}
\def\uZ{\underline{\Z}}

\def\dist{{\rm dist}}

\def\Lip{{\rm Lip}}

\def\rank{{\rm rank}}

\def\distCD{d_{C}} 

\def\ran{{\rm range}}
\def\X0{X_0}

\def\eqnn{\begin{eqnarray*}}
\def\eeqnn{\end{eqnarray*}}
\def\eqn{\begin{eqnarray}}
\def\eeqn{\end{eqnarray}}

\def\prf{\begin{proof}}
\def\endprf{\end{proof}}

\theoremstyle{plain}
\newtheorem{theorem}{Theorem}[section]
\newtheorem{definition}[theorem]{Definition}
\newtheorem{proposition}[theorem]{Proposition}

\newtheorem{lemma}[theorem]{Lemma}

\newtheorem{remark}[theorem]{Remark}

\numberwithin{equation}{section}

\begin{document} 

\title[Generalization in neural networks]
{Architecture independent generalization bounds for overparametrized deep ReLU networks}
 
\author[A. Bapu]{Anandatheertha Bapu}
\address[A. Bapu]{Department of Mathematics, University of Texas at Austin, USA}
\email{anandatheerthabapu@gmail.com}

\author[T. Chen]{Thomas Chen}
\address[T. Chen]{Department of Mathematics, University of Texas at Austin, USA}
\email{tc@math.utexas.edu}  

\author[C.-K. Chien]{Chun-Kai Kevin Chien}
\address[C.-K. Chien]{Mathematisches Institut, Universit\"at M\"unster}
\email{ck.chien@uni-muenster.de}  

\author[P. Mu\~{n}oz Ewald]{Patricia Mu\~{n}oz Ewald}
\address[P.M. Ewald]{Department of Mathematics, University of Texas at Austin, USA}
\email{ewald@utexas.edu}  

\author[A.G. Moore]{Andrew G. Moore}
\address[A.G. Moore]{Department of Mathematics, University of Texas at Austin, USA}
\email{agmoore@utexas.edu}  

\begin{abstract}
We prove that overparametrized neural networks are able to generalize with a test error
that is independent of the level of overparametrization, and independent of the
Vapnik-Chervonenkis (VC) dimension. We prove explicit bounds that only depend on the
metric geometry of the test and training sets, on the regularity properties of the
activation function, and on the operator norms of the weights and norms of biases. For
overparametrized deep ReLU networks with a training sample size bounded by the input space
dimension, we explicitly construct zero loss minimizers without use of gradient descent,
and prove a uniform generalization bound that is independent of the network architecture.
We perform computational experiments of our theoretical results with MNIST, and obtain agreement
with the true test error within a $22\%$ margin on average.
\end{abstract}
\maketitle

\section{Introduction and main results}

We consider a neural network with input space $\cL_0\cong\R^{M_0}$, hidden layers
$\cL_\ell\cong \R^{M_\ell}$ with $\ell=1,\dots,L$, and output layer $\cL_{L+1}\cong \R^Q$.
Given $x^{(0)}\in\cL_0$, we recursively define 
\eqn\label{eq-layermap-1-0}	
	x^{(\ell)} = \sigma(W_\ell x^{(\ell-1)}+b_\ell)
\eeqn  
with weight matrix $W_\ell\in\R^{M_\ell\times M_{\ell-1}}$ and bias vector $b_\ell\in \R^{M_\ell}$ for $\ell=1,\dots,L$. 
We assume that the activation function $\sigma:\R^{M}\rightarrow {\rm ran}(\sigma)\subset\R^M$ is Lipschitz continuous, and satisfies a growth bound
\eqn\label{eq-sigma-bd-1-0}
	|\sigma(x)| \leq a_0+a_1 |x|
\eeqn
for some constants $a_0\geq0$, and $a_1>0$.
In the output layer,
\eqn	
	x^{(L+1)} = W_{L+1} x^{(L)}+b_{L+1}
\eeqn 
we assume no activation to be present.

Let $\uZ\in\R^K$ denote the vector of parameters, containing the components of all weights and biases, so that
\eqn 
	K=\sum_{\ell=1}^{L+1}(M_\ell M_{\ell-1}+M_\ell) \,.
\eeqn 	
We will assume architectures with $M_0\geq M_1 \geq \cdots \geq M_L \geq Q$. 
For every $\uZ\in\R^K$, the network defines the map 
\eqn 
	f_{\uZ}:\cL_0\rightarrow \cL_{L+1}
	\;\;\;,\;\;\;
	x^{(0)}\mapsto x^{(L+1)}=f_{\uZ}(x^{(0)})
\eeqn 
from input to output layer. 

We consider a problem in supervised learning, 
and assume
to be given an {\em a priori sample set }
\eqn 
	\cS:= \{(x_{i}^{(0)},y_i)\}_{i=1}^{N}
\eeqn 
where $x_{i}^{(0)}\in\cL_0\cong\R^{M_0}$ and $y_i\in\cL_{L+1}\cong\R^Q$, with $M_0 \geq Q
\geq 1$.

We extract two subsets of $\cS$, typically disjoint, one of which we refer to as the {\em training set}
\eqn 
	\cS^{train} := \{(x_{i_\alpha}^{(0)},y_{i_\alpha})\}_{\alpha=1}^{n} \,,
\eeqn
and 
the other as the {\em test set}
\eqn 
	\cS^{test} :=  \{(x_{i_\beta}^{(0)},y_{i_\beta})\}_{\beta=1}^{m} \,.
\eeqn
For notational brevity, we will denote the elements of these sets by
\eqn 
	\cS^{train} = \{(x_{i}^{(0)},y_i)\}_{i=1}^{n}
\eeqn
and
\eqn 
	\cS^{test} = \{(\widetilde x_{j}^{(0)},\widetilde y_j)\}_{j=1}^{m} \,,
\eeqn
upon appropriate relabeling.

Then, the network is trained by minimizing the cost (or loss) function
\eqn 
	\cC := \frac{1}{n}\sum_{i=1}^{n} |f_{\uZ}(x_{i}^{(0)})-y_i|^2
\eeqn 
with respect to $\uZ$. Let $\uZ_*$ denote a global minimizer of $\cC$ with fixed training set $\cS^{train}$. Then,  
\eqn
	\cE_{\uZ_*}^{train} := 
	\frac{1}{n}\sum_{i=1}^{n} |f_{\uZ_*}(x_{i}^{(0)})-y_i|^2
\eeqn  
is the {\em training error}. It is zero if zero loss is attainable.

The test error is determined by evaluating $\cS^{test}$ on the trained network,
\eqn 
	\cE_{\uZ_*}^{test} := 
	\frac{1}{m}\sum_{j=1}^{m} |f_{\uZ_*}(\widetilde x_{j}^{(0)})-\widetilde y_j|^2 \,.
\eeqn 
In the work at hand, we will not assume that the probability distributions of $\cS^{train}$ and $\cS^{test}$ are a priori known, see the remarks in Section \ref{remarks-1-0}.

\begin{definition}
	We call a network {\em strongly overparametrized} if the training sample size is bounded by the dimension of input space, $n\leq M_0$.
\end{definition}

We note that this is a reasonable assumption in contexts where good quality labeled
data is difficult to acquire, such as in medical and hyperspectral imaging; see e.g.
\cite{wang2023advances, schafer2024overcoming, groger2025review}.

\subsection{A priori bound}
The difference between $\cE_{\uZ_*}^{test}$ and $\cE_{\uZ_*}^{train}$ is accounted for by
the generalization error. We obtain the following a priori bounds in Proposition \ref{thm-generr-1-0}, which we strengthen significantly in Theorem \ref{thm-main-zeroloss-0-1}. In Theorem \ref{thm-C0-ReLU-1-0}, we further strengthen the latter to prove to be architecture independent for the specific case of overparametrized deep ReLU networks.

Given $\cS^{train}$ and $\cS^{test}$,
the generalization error satisfies
\eqn\label{eq-cE-diff-1-0}
		|\cE_{\uZ_*}^{test} -  \cE_{\uZ_*}^{train}|
		=
		\cD[\cS^{test} ,\cS^{train} ] \,,
\eeqn 
where we refer to 
\eqn 
	\cD[\cS^{test} ,\cS^{train} ] :=
	\frac{1}{n m}\Big|\sum_{i=1}^{n}\sum_{j=1}^{m}
	\Big( 
	|f_{\uZ_*}(x_{i}^{(0)})-y_i|^2 -
	|f_{\uZ_*}(\widetilde x_{j}^{(0)})-\widetilde y_j|^2
	\Big)\Big|
\eeqn 
as the loss discrepancy between $\cS^{test}$, $\cS^{train}$.
In particular, 
\eqn
	\cD[\cS^{test} ,\cS^{train} ] =0
	\;\;\;{\text if}\;\;\;\cS^{test}=\cS^{train} \,,
\eeqn 
and more generally, it vanishes whenever $\cE_{\uZ_*}^{test} = \cE_{\uZ_*}^{train}$.

\begin{proposition}
\label{thm-generr-1-0}
The loss discrepancy satisfies the a priori bound
\eqn\label{eq-cD-apriori-1-0}
	\cD[\cS^{test} ,\cS^{train} ] 
	\leq C_0 (1+R) \; {\rm diam}(\cS^{test}\cup\cS^{train})
\eeqn 
where $R$ is the radius of the smallest ball centered at the origin in $\R^{M_0}\times\R^Q$ containing $\cS^{test}\cup\cS^{train}$, and ${\rm diam}(\cS^{test}\cup\cS^{train})$ is the diameter of $\cS^{test}\cup\cS^{train}$. 

The constant  
\eqn\label{eq-C0-bd-1-0-0}
	C_0 = C_0(a_0,a_1,c_\sigma,\{\|W_\ell^*\|_{op},|b_\ell^*|\}_{\ell=1}^{L+1})
\eeqn 
depends on the operator norms of the trained weights and the Euclidean norms of the trained biases. Moreover, it depends on the parameters $a_0,a_1$ controlling the growth of $\sigma$ in \eqref{eq-sigma-bd-1-0}, and its Lipschitz constant $c_\sigma$.
\end{proposition}

\subsubsection{Remarks}

Using $b_1$, the origin in input space can be translated to minimize $R$. By Jung's theorem, one then obtains $R \leq \sqrt{\frac{M_0}{2(M_0+1)}}\,{\rm diam}(\cS^{test}\cup\cS^{train})$.

The a priori bound \eqref{eq-C0-bd-1-0-0} on $C_0$ involves the product of operator norms of all weights, and can generically be very large.
For a related result, see \cite{neytomsre-1}.

\subsection{Generalization bound}

We can improve this result significantly in Theorem \ref{thm-main-zeroloss-0-1} if we
bound $\mathcal{E}_{\underline{\theta}_{*}}^{test}$ separately instead of
$|\mathcal{E}_{\underline{\theta}_{*}}^{test} -
\mathcal{E}_{\underline{\theta}_{*}}^{train}|$,
and even much further for ReLU networks in Theorem
\ref{thm-C0-ReLU-1-0}, where we prove bounds that do not depend on the depth of the
network nor on the widths of its hidden layers.


\begin{theorem}
\label{thm-main-zeroloss-0-1}
Let $g_{\uZ_*}:\R^{M_1}\rightarrow \R^Q$ denote the map from the first hidden layer to the output layer for the trained network, so that
\eqn 
	f_{\uZ_*}(x) = g_{\uZ_*}(W_1^*x+b_1^*)\,,
\eeqn 
and let $\Lip(g_{\uZ_*})$ denote its Lipschitz constant.
Let
\eqn 
	\cS^{train}_{W_1^*}&:=&\{(W_1^* x,y)\;|\;(x,y)\in \cS^{train}\}
	\nonumber\\
	\cS^{test}_{W_1^*}&:=&\{(W_1^*  \widetilde x,\widetilde y)\;|\;
	(\widetilde x,\widetilde y)\in \cS^{test}\}\,.
\eeqn  
Then, defining 
\eqn\label{eq-CuZ-Lip-1-0}
	C_{\uZ_*} := \max\{1,\Lip(g_{\uZ_*})\} \,,
\eeqn 
it follows that
\eqn\label{eq-generr-zero-1-0} 
	\cE_{\uZ_*}^{test} \leq  C_{\uZ_*}^2 \; 
	\distCD^2(\cS^{test}_{W_1^*}\,|\,\cS^{train}_{W_1^*} ) \,
\eeqn 
when $\mathcal{E}_{\underline{\theta}_{*}}^{train} = 0$, and in general
\eqn
	\cE_{\uZ_*}^{test} \leq 
    (1 + \delta)C_{\uZ_*}^2 \distCD^2(\cS^{test}_{W_1^* }\,|\,
	\cS^{train}_{W_1^* }) + \left(1+ \frac{1}{\delta}\right)
    \mathcal{E}_{\underline{\theta}_{*}}^{testNN}
\eeqn
for any $\delta > 0$, where 
\begin{align*}
    \mathcal{E}_{\underline{\theta}_{*}}^{testNN} = \frac{1}{m}\sum_{j=1}^{m} 
    r(x_{i(j)}^{(0)}, y_{i(j)})^{2} 
\end{align*}
is the average training error for points in $\mathcal{S}^{train}$ which are nearest
neighbors to points in $\mathcal{S}^{test}$.
These bounds hold with respect to the unidirectional Chamfer pseudodistance for point clouds,
\eqn\label{eq-distCD-def-1-0}
	\distCD(\cS^{test}_{W_1^*}\,|\,
	\cS^{train}_{W_1^*}):=\left(\frac1m
	\sum_{(\widetilde x,\widetilde y)\in\cS^{test}}
	\min_{(x,y)\in\cS^{train} }
	\big(|W_1^*(\widetilde x-x)|+|\widetilde y-y|\big)^2\right)^{\frac12}\,.
\eeqn
In particular, if zero loss training is achieved, the trained weight $W_1^*$ acts projectively, and is supported on the span of the training data.
\end{theorem}

For strongly overparametrized deep ReLU networks, zero training loss is attainable with
explicitly constructible minimizers, and $C_{\uZ_*}$ in \eqref{eq-generr-zero-1-0} is
independent of the network architecture, see Theorem \ref{thm-C0-ReLU-1-0}. In Section
\ref{experiments}, we show empirically for randomly initialized ReLU networks trained on the MNIST
data set that, for very small $n$, the bound \eqref{eq-generr-zero-1-0} does not change
significantly with depth.


\subsubsection{Remarks}
The unidirectional Chamfer pseudodistance does {\em not} satisfy the axioms of a metric. Along with its symmetrized version, it is widely used in computer vision, \cite{niesansararacre}. It measures the approximability of $\cS^{test}_{W_1^* }$ by subsets of $\cS^{train}_{W_1^* }$, but not vice versa. In particular, $\distCD(\cS^{test}_{W_1^* }\,|\,\cS^{train}_{W_1^* })=0$ implies that $\cS^{test}_{W_1^* }\subseteq\cS^{train}_{W_1^* }$. 

For Theorem \ref{thm-main-zeroloss-0-1}, it is {\em not necessary} for $\cS^{test}_{W_1^*}$ and $\cS^{train}_{W_1^* }$ to be identically distributed.  

We also note that in
\eqref{eq-distCD-def-1-0}, the trained weight $W_1^*$ acts projectively, and is supported on the span of the training data if zero training loss achieved. This means that 
\eqn
    W_1^*(\widetilde x-x)=W_1^*(P_{X^{(0)}}\widetilde x-x)
	\;\;\;{\rm and}\;\;\;
    P_{X^{(0)}}x=x
\eeqn 
for all $\widetilde x\in\cS^{test}$ and $x\in\cS^{train}$, where $X^{(0)}$ is a matrix
with columns formed by the $x^{(0)}$ vectors in $\mathcal{S}^{train}$, and $P_{X^{(0)}} = X^{(0)}
(X^{(0)})^{+}=P_{X^{(0)}}^T$ projects orthogonally onto $\ran(X^{(0)})$.
Here, $X_0^+$ is the generalized inverse, and $X_0^+=(X_0^T X_0)^{-1}X_0^T$ when $X_0$ has full rank.
In other words, only the component of $x$ in the linear span of the training data are relevant for the generalization error. 
Moreover, we point out that for all training data in the set 
\eqn 
	\cS^{train}_{W_1^*} = \{(W_1^*x,y)|(x,y)\in \cS^{train}\}
\eeqn 
we have $W_1^*x=(W_1^*X^{(0)}) (X^{(0)})^+x$ with $W_1^*X^{(0)}=Y-B_1^*$, and the vectors
$(X^{(0)})^+x$ are coordinate unit vectors of the form $e_\ell=(0,\dots,0,1,0,\dots,0)$
($\ell$-th entry is 1).

This means that the trained network projects the test data set onto the span of the training data set, which equals $\ran(X^{(0)})$ in $\R^{M_0}$. Therefore, the components of test data orthogonal to the span of the training data do not enter the generalization bound. We interpret this a key reason for the absence of overfitting.



\subsection{Uniform generalization bounds for zero loss ReLU}

In Theorem \ref{thm-C0-ReLU-1-0}, below, we determine an explicit bound on the constant
$C_0$  for the important special case of strongly overparametrized deep ReLU networks. We
present the construction of zero training loss minimizers without invoking gradient
descent, and prove that the product of operator norms of weights in $C_0$ collapses,
allowing for a {\em uniform} bound {\em independent of the network architecture}.

For notational convenience, we define the matrices associated to $\cS^{train}$,
\eqn\label{eq-Xell-def-1-0}
	X^{(\ell)} := [x_1^{(\ell)}\cdots x_n^{(\ell)}] 
	\;\;\; \in \; \R^{M_\ell\times n}
\eeqn  
and 
\eqn\label{eq-Bell-def-1-0}
	B_\ell := [b_\ell\cdots b_\ell ]
	\;\;\; \in \; \R^{M_\ell\times n}
\eeqn 
for $\ell=1\cdots,L+1$, with $M_{L+1}=Q$. Moreover,
\eqn 
	Y:=[y_1\cdots y_n] \;\;\;\in\;\R^{Q\times n}
\eeqn 
is the matrix of reference output vectors $y_i\in\cL_{L+1}\cong\R^Q$ in output space.

We obtain the following bound for strongly overparametrized ReLU networks with non-increasing layer dimensions, yielding a uniform generalization bound which only depends on the training input data and labels. In particular, it is independent of the layer dimensions and of the number of layers.

\begin{theorem}\label{thm-C0-ReLU-1-0}
	Consider a strongly overparametrized network with $M_0\geq M_1\geq \cdots \geq M_L\geq Q$ and training sample size $n\leq M_0$. Assume that the matrix of training inputs $X_0\equiv X^{(0)}$ has full rank $n$, and that $\sigma$ is ReLU. Then, zero loss training is attainable with explicitly constructible minimizer.

%
The zero training loss minimizer is explicitly given by
\eqn 
	W_1^* &=& 
	\left[
	\begin{array}{c}
	(Y - B_{L+1}^*)
	(X^{(0)})^+
	\\
	0_{(M_{1}-Q)\times M_0}
	\end{array}
	\right]
	\nonumber\\
	W_\ell^*&=&[\1_{M_\ell\times M_\ell} \;\;0_{(M_{\ell-1}-M_\ell)\times M_L}]\;\;\;,\;\;\ell=2,\dots,L
	\nonumber\\
	W_{L+1}^* &=& [\1_{Q\times Q} \;\;0_{Q\times (M_{L}-Q)}]
	\nonumber\\
	b_\ell^*&=&0\;\;\;,\;\;\ell=1,\dots,L
	\nonumber\\
	b_{L+1}^*&=&-\alpha(1,\dots,1)^T 
	\;\;\;,\;\;
	\alpha =  \max_{i,j}|(Y_{ij})_-|  \,.
\eeqn
Therefore, $\|W_\ell^*\|_{op}=1$ for all $\ell=2,\dots,L$, and hence, the constants in \eqref{eq-CuZ-Lip-1-0} have the values $\Lip(g_{\uZ_*})=1$ and $C_{\uZ_*}=1$.

Consequently, 
	the generalization bound
\eqn\label{eq-generr-zero-1-1} 
	\cE_{\uZ_*}^{test} &<& 
	\frac1m
	\sum_{(\widetilde x,\widetilde y)\in\cS^{test}}
	\min_{(x,y)\in\cS^{train} }
	\big(|W_1^*(\widetilde x-x)|+|\widetilde y-y|\big)^2
	\nonumber\\
	&=&  
	\; \distCD^2(\cS^{train}_{W_1^*} \, | \, \cS^{test}_{W_1^*}) \,.
\eeqn 
follows from \eqref{eq-generr-zero-1-0}.
\end{theorem}

\subsubsection{Remarks}
Here, $\alpha$ is the modulus of the most negative coordinate among all labels $y_j$.
We remark that in various classification problems, $y_i$ are chosen as unit vectors, with all components non-negative, whereby $\alpha=0$.


With straightforward modifications, we expect that a similar result can be proven for the same architecture, but for an activation given by a diffeomorphism 
\eqn
	\sigma:\R^M\rightarrow\R^M_+
\eeqn
(for instance, a suitable mollification of ReLU), as studied in \cite{chemoo-1}.

Notably, the zero loss minimizers are not unique,
see also \cite{cheewa-5,chemoo-1,coo,lidingsun,zha}.

\section{Comparison with probabilistic bounds}

In this section, we compare our results with standard approaches developed in Statistical Learning Theory. 

\subsection{Probabilistic setting}
\label{remarks-1-0}
If it is assumed that the probability distributions for the training set $\cS^{train}$  and test set $\cS^{test}$ are known, and determined by probability measures $\mu_{train}$ and $\mu_{test}$ on $\R^{M_0}\times\R^Q$, respectively, then
\eqn 
	\cE_{\uZ_*}^{train} &= &
	\int d\mu_{train}( x, y) |f_{\uZ_*}( x)- y|^2  
	\nonumber\\
	\cE_{\uZ_*}^{test} &= &
	\int d\mu_{test}(\widetilde x,\widetilde y) |f_{\uZ_*}(\widetilde x)-\widetilde y|^2 \,.
\eeqn 
The generalization error is usually analyzed by controlling the probability of the event that their difference is small.
Probabilistic bounds on the generalization error in this context are studied extensively in the literature, see for instance \cite{barfostel,nagkol-1,neybhomcasre-1,zhabenharrecvin-1} and the references therein. 
We point out two crucial difficulties with this setting:
\begin{enumerate}
	\item In practice, the probability distributions for the test and training sets are typically not  a priori known.
	\item It is usually assumed that $\cS^{train}$  and $\cS^{test}$ are sampled from the same set, and distributed identically,  which suggests proximity of $\cE_{\uZ_*}^{train}$ and $\cE_{\uZ_*}^{test}$. 
\end{enumerate}

Point (2) introduces an extra constraint. In practice, $\mu_{train}$ and $\mu_{test}$ might not be identically distributed. 
	For example, a network could be trained to distinguish apples from oranges with a training set $\cS^{train}$ from country A. It should then be expected to have the ability to distinguish apples and oranges in a set $\cS^{test}$ originating from country B that is not identically distributed to $\cS^{train}$.


Because of point (1), concrete implementations commonly employ a similar, practicable
version of generalization error as in Proposition \ref{thm-generr-1-0}. We point out that
Proposition \ref{thm-generr-1-0} holds for arbitrary but fixed $\cS^{train}$ and
$\cS^{test}$, without any assumptions on their probability distributions. In Theorem
\ref{thm-C0-ReLU-1-0}, we will prove for strongly overparametrized ReLU networks that zero
loss is attainable and that the generalization error then is independent of the network
architecture, only depending on how well $\cS^{test}$ can be approximated with subsets of
$\cS^{train}$ of equal cardinality.

\subsection{Bounds based on the Vapnik-Chervonenkis dimension}

The bound on the generalization error most widely used in the literature is an expression involving the Vapnik-Chervonenkis (VC) dimension $d_{VC}$ of the network. 
The statement is that with probability at least $1 -\delta$,
\eqn\label{eq-VCbound-1-0} 
	\cE_{\uZ_*}^{test} \leq \cE_{\uZ_*}^{train}
	+ O\Big( \sqrt{ \frac{d_{VC}}{n} } + \sqrt{\frac{\log(1/\delta)}{n}}\Big)
\eeqn 
holds for {\em  identically distributed} $\cS^{train}$ and $\cS^{test}$. $d_{VC}$ accounts for the maximum size of a point set allowing for all of its subsets to be distinguishable by the network. 

\subsubsection{Overparametrized networks}
\label{ssec-overpar-1-0}
For neural networks, $d_{VC}$ has been determined in \cite{barmaa}. In particular, $d_{VC}$ typically increases with the level of overparametrization of the problem; that is, with the amount by which $K$ exceeds the overall output dimension $nQ$. 

The bound \eqref{eq-VCbound-1-0} deteriorates with the increase of overparametrization and thus of $d_{VC}$; however, computational evidence shows that the generalization error remains bounded. This conundrum is a key open problem in Machine Learning, see the discussion of the double descent phenomenon in \cite{bel,belhsumaman-1}. The error term in \eqref{eq-VCbound-1-0}  decreases as the number of training data $n$ increases, but this in turn lowers the level of overparametrization; in particular, in the limit $n\rightarrow\infty$, the problem becomes underparametrized. Therefore, the bound \eqref{eq-VCbound-1-0} is not effective in the overparametrized regime where control of the generalization error is most desired.

In this paper, we offer the following explanation to this phenomenon: The VC dimension describes the capacity of the network to accommodate any given set of input data. However, the bound does not account for the specific data determining the problem, and is independent of both $\cS^{test}$ and $\cS^{train}$. In Theorem \ref{thm-C0-ReLU-1-0} of this paper, we verify that the true generalization error for strongly overparametrized deep ReLU networks is determined by metric properties of the data distribution independent of the parameter count, and of the network architecture. For a discussion related to ours, we refer to \cite{nagkol-1}.

Some intuition about the situation at hand can perhaps be gained from the following analogy: 1 liter of fluid can be filled into a 1 liter bottle, a 2 liter bottle, or a 50 liter bottle; however, in all cases, the amount of fluid remains the same. The size of the bottle corresponds to the VC dimension of the network, it grows as the overparametrization level increases. The amount of fluid corresponds to the test and training data sets, which are independent of the network architecture.

\subsubsection{Zero loss in underparametrized regime}

For generic training data, underparametrized networks are not able to admit zero loss. This is because the map from parameter space to output space (for given training data) is an embedding, and the zero loss minimizer is generically not contained in its range. However, training with zero loss in underparametrized networks is possible if $\cS^{train}$ is sufficiently clustered, and therefore not generic, \cite{cheewa-3,chemoo-1}. 

In \cite{cheewa-2,cheewa-4}, zero loss minimizers are explicitly constructed for underparametrized deep ReLU networks with sufficiently clustered training data. It is assumed that there are $Q$ disjoint clusters corresponding to $Q$ equivalence classes, and the depth of the network is required to be at least $Q$. The weights and biases of the trained network are explicitly determined in \cite{cheewa-2,cheewa-4}, without invoking gradient descent. Each layer acts by collapsing one of the clusters of training data to a point, and in the output layer, the resulting $Q$ disjoint points are matched to the vectors $y_i$, $i=1,\dots,Q$ that label the classes. Thereby, zero loss training is achieved. 

A crucial question arises: The generalization error in this network is controlled by the bound \eqref{eq-VCbound-1-0}, based on its VC dimension. The training error is zero, hence
\eqn\label{eq-VCbound-1-1} 
	\cE_{\uZ_*}^{test} \leq  O\Big( \sqrt{ \frac{d_{VC}}{n} } + \sqrt{\frac{\log(1/\delta)}{n}}\Big) \,.
\eeqn  
Therefore, taking the limit $n\rightarrow\infty$ of infinite training sample size, \eqref{eq-VCbound-1-1} predicts that the test error must be zero, too.

However, with the construction presented in \cite{cheewa-2,cheewa-4}, it is easy to create full measure sets of test data in the input layer (sufficiently separated from the training clusters) that do {\em not} produce zero loss. This might seem like a contradiction.

However, there is no contradiction; the resolution of this conundrum is that for \eqref{eq-VCbound-1-0} to hold, it is required that the test sample set $\cS^{test}$ and training data set $\cS^{train}$ are {\em identically distributed}. Therefore, to apply \eqref{eq-VCbound-1-1}, test samples are only admissible if they are clustered with identical distribution as the training data; but then, the trained network maps them to output vectors with zero loss, as well.  

To summarize, the discussion in Section \ref{ssec-overpar-1-0} suggests that the effectiveness of \eqref{eq-VCbound-1-0} in the overparametrized regime is limited. On the other hand, \eqref{eq-VCbound-1-0} is effective in the underparametrized regime, where identical distribution of test and sample data is a necessary condition. If the latter is not satisfied, one can construct examples violating \eqref{eq-VCbound-1-0}.

\section{Experiments} \label{experiments}

To verify our theoretical bound \eqref{eq-generr-zero-1-0} in Theorem
\ref{thm-main-zeroloss-0-1}, we trained shallow feedforward neural networks with layer
dimensions $(784, 784, 10)$, no bias vectors, and ReLU activation to classify MNIST digits in three
different ways: First, we constructed the zero loss minimizers in Theorem
\ref{thm-C0-ReLU-1-0}, abbreviated ZL below. We used these constructive minimizers as 
the initialization to train new networks (TFZL, ``trained from zero loss''), to ascertain that they are stable under gradient descent. Finally, we
trained networks from random initialization (RI).

All training was done using stochastic gradient descent with 70 epochs
and learning rate $0.1$.
The training samples are picked randomly out of MNIST, under the condition that all
digits $\{0,1,\dots,9\}$ are represented; samples violating this requirement are rejected.
The code is made available on the website \url{https://github.com/patriciaewald/BCCEM}.

\begin{figure}[h]
    \centering
    \includegraphics[width=0.49\textwidth]{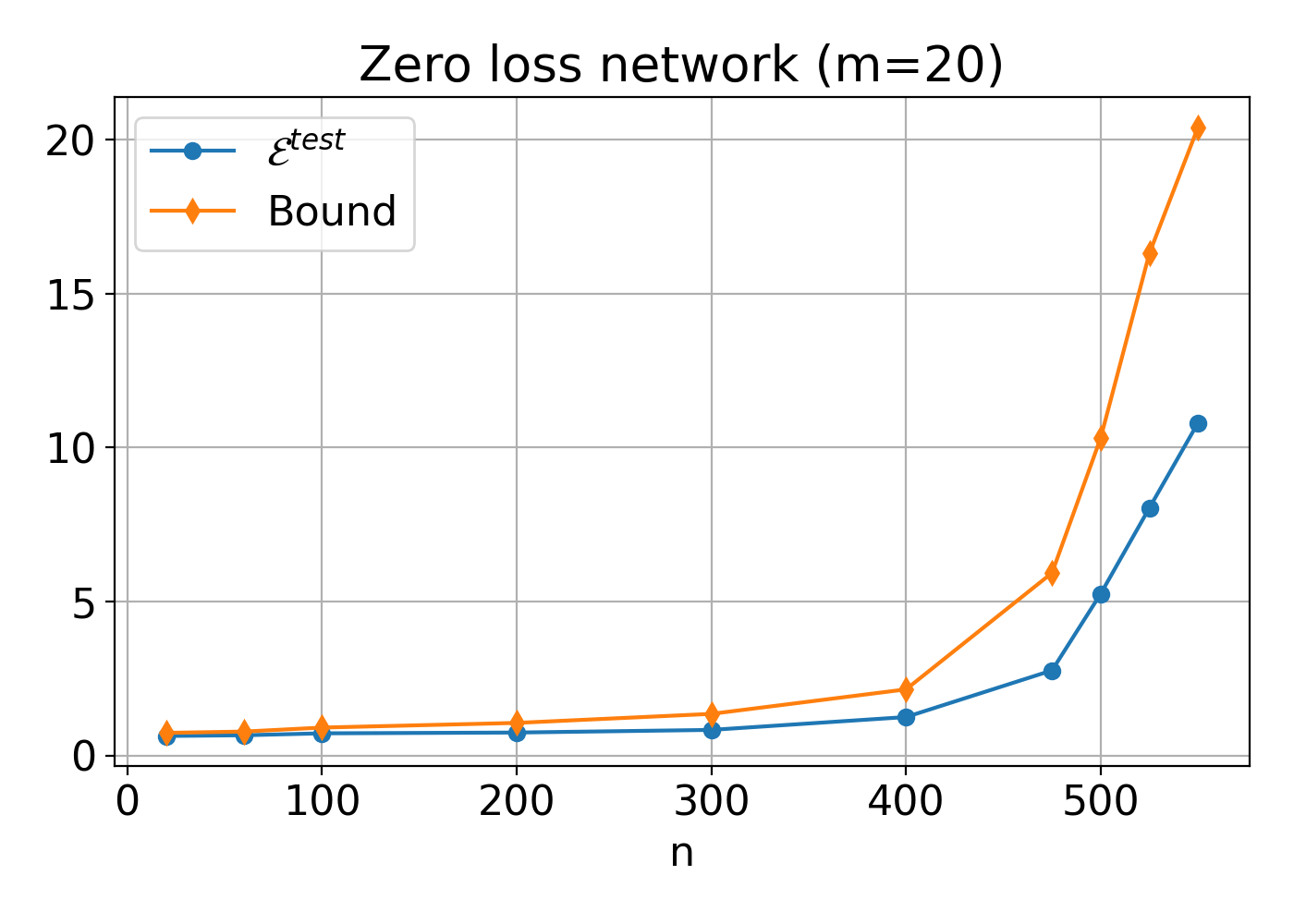}
    \includegraphics[width=0.5\textwidth]{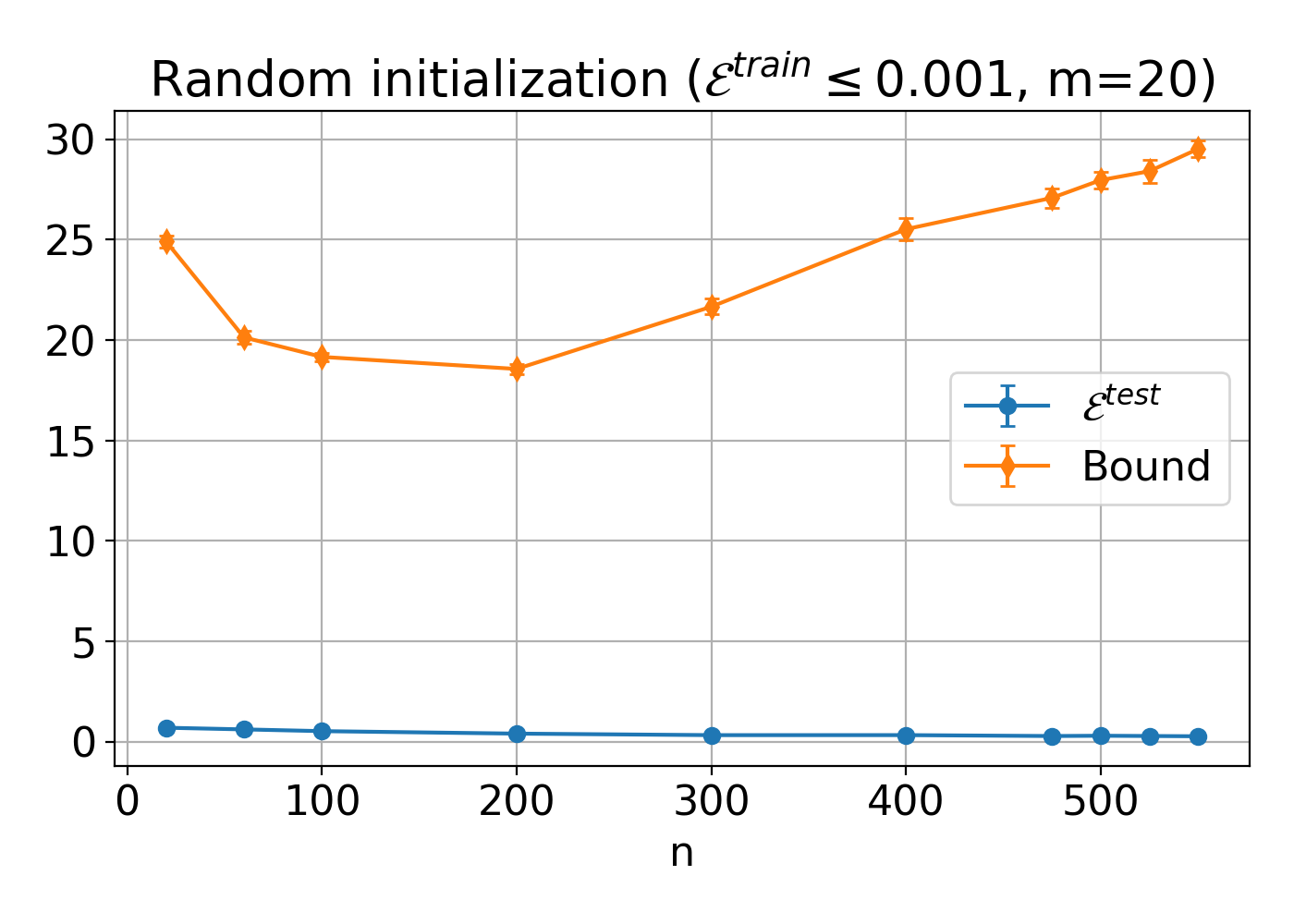}
    \caption{Each plot shows the test error $\mathcal{E}^{test}$ and the bound
    \eqref{exp-bound} for the constructed or trained shallow networks. The plot for TFZL is very
    similar to the one for ZL (left). }
    \label{Etest-bound-plots}
\end{figure}

\begin{figure}
    \centering
    \includegraphics[width=0.7\textwidth]{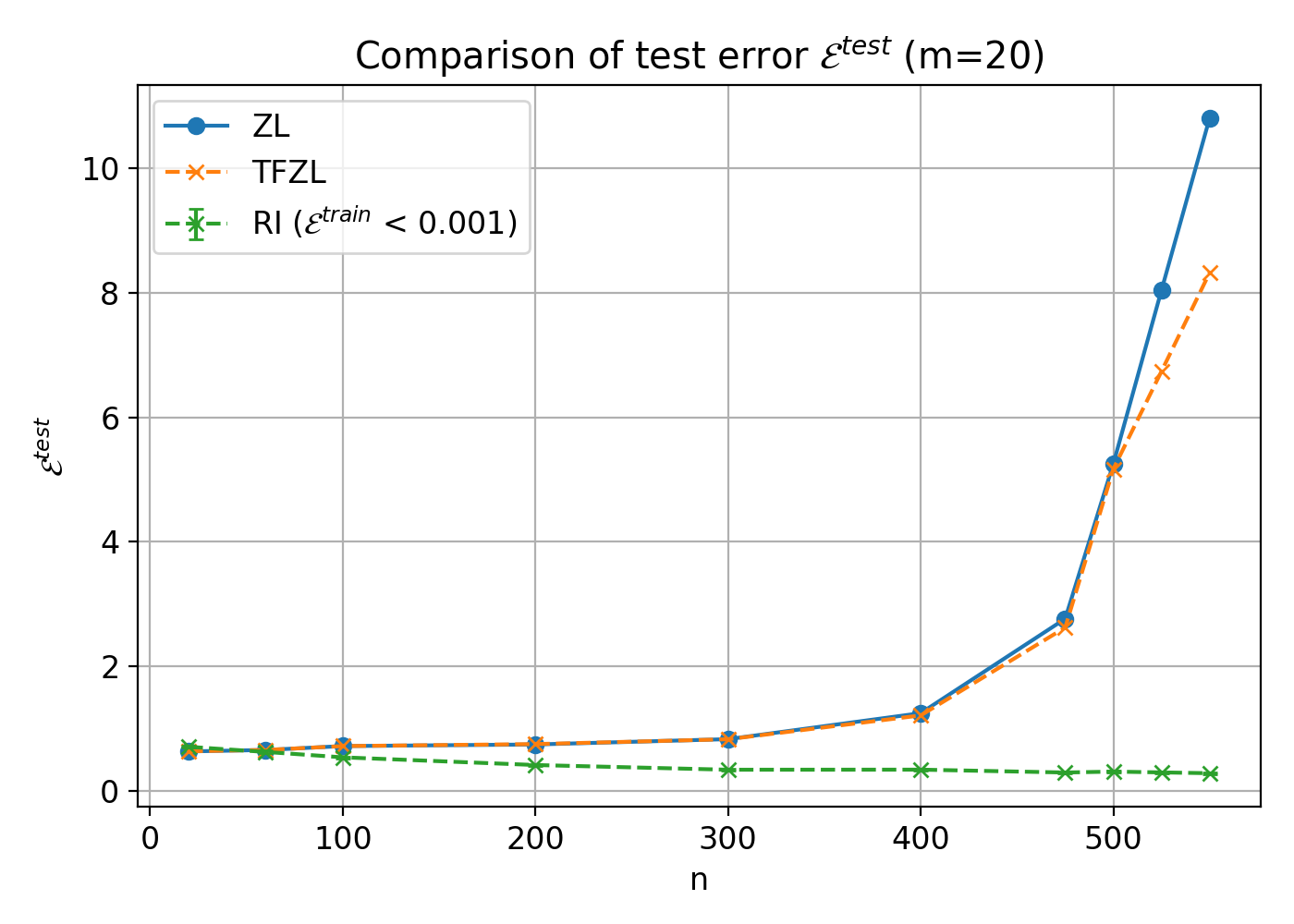}
    \caption{This plot shows the resulting test error for a fixed test set and several
        shallow networks trained with training sets of different sizes $n$. }
    \label{compare-Etest}
\end{figure}

In Figures \ref{Etest-bound-plots} and \ref{compare-Etest} we have plotted the resulting
test error $\mathcal{E}^{test}$ and the following upper bound on \eqref{eq-generr-zero-1-0}
\begin{align}
    \label{exp-bound}
    \text{Bound } = \max\{1, \Pi_{\ell =2}^{L+1}||W_{\ell }||_{op} \} ^{2} d_{C}^{2}(\mathcal{S}^{test}_{W_1} |
    \mathcal{S}_{W_1}^{train}),
\end{align}
for a ReLU network with $L$ hidden layers, as a function of
the size of the training data set. For the shallow neural networks, the product of norms
simplifies to $||W_2||_{op}$. For each value of  $n$, there is one instance of
ZL and TFZL, whereas the plotted values for RI are averaged over several random
initialiations. The test set was kept fixed, with $m=20$ samples. 
Note that $n$ was capped at $550$ to stay within the assumptions of Theorem
\ref{thm-C0-ReLU-1-0}, whereby the data matrix $X_0$ needs to be full-rank.

Notably, our theoretical generalization bound yields good agreement with
the constructed ZL network in the strongly overparametrized regime where network complexity
based bounds are not effective \cite{bel}. While the agreement with the randomly
initialized networks is not as good, the bound for ZL represents
an upper bound on the best generalization error:
\begin{align}
    \min_{\underline{\theta}} \mathcal{E}_{\underline{\theta}}^{test} \leq \text{Bound for
    ZL}.
\end{align}
This is particularly valuable in the data scarcity context: 
In Figure \ref{compare-Etest} we observe that 
as $n$ gets smaller (or as the system gets more overparametrized), the average test error for the
randomly initialized networks gets closer to, and indeed surpasses, the test error for the
explicitly constructed network (ZL).

\begin{table}[h]
    \centering
    \begin{tabular}{lccc}
         & Min & Max & Average \\
        \hline
      	ZL   & 1.15  & 1.42  & 1.22 \\
		TFZL & 1.15  & 1.43  & 1.22 \\
		RI   & 29.21 & 48.09 & 34.51 \\
    \end{tabular}
    \caption{For each network, we computed the
    ratio $\gamma = \frac{Bound}{\mathcal{E}^{test}}$. In this table, we report the
    minimum, maximum, and average values of $\gamma$ for each type of trained network, in
    the range $n \in [20, 200]$.}
    \label{table}
\end{table}

\begin{figure}[h]
    \centering
    \includegraphics[width=0.41\textwidth]{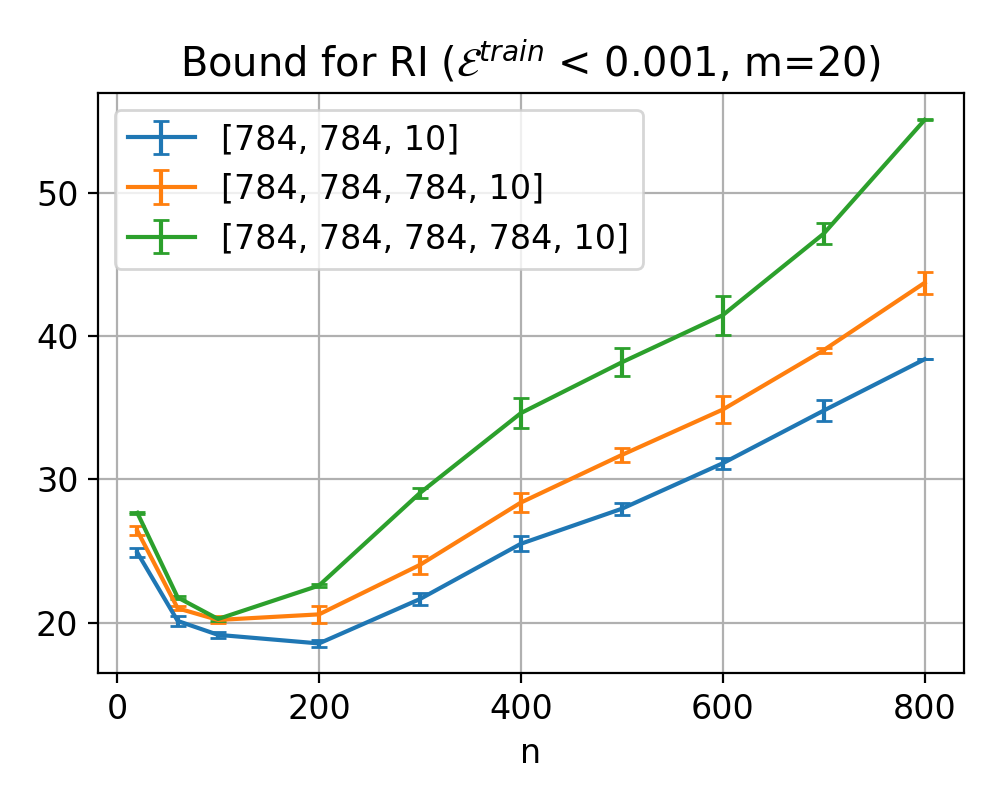}
    \includegraphics[width=0.58\textwidth]{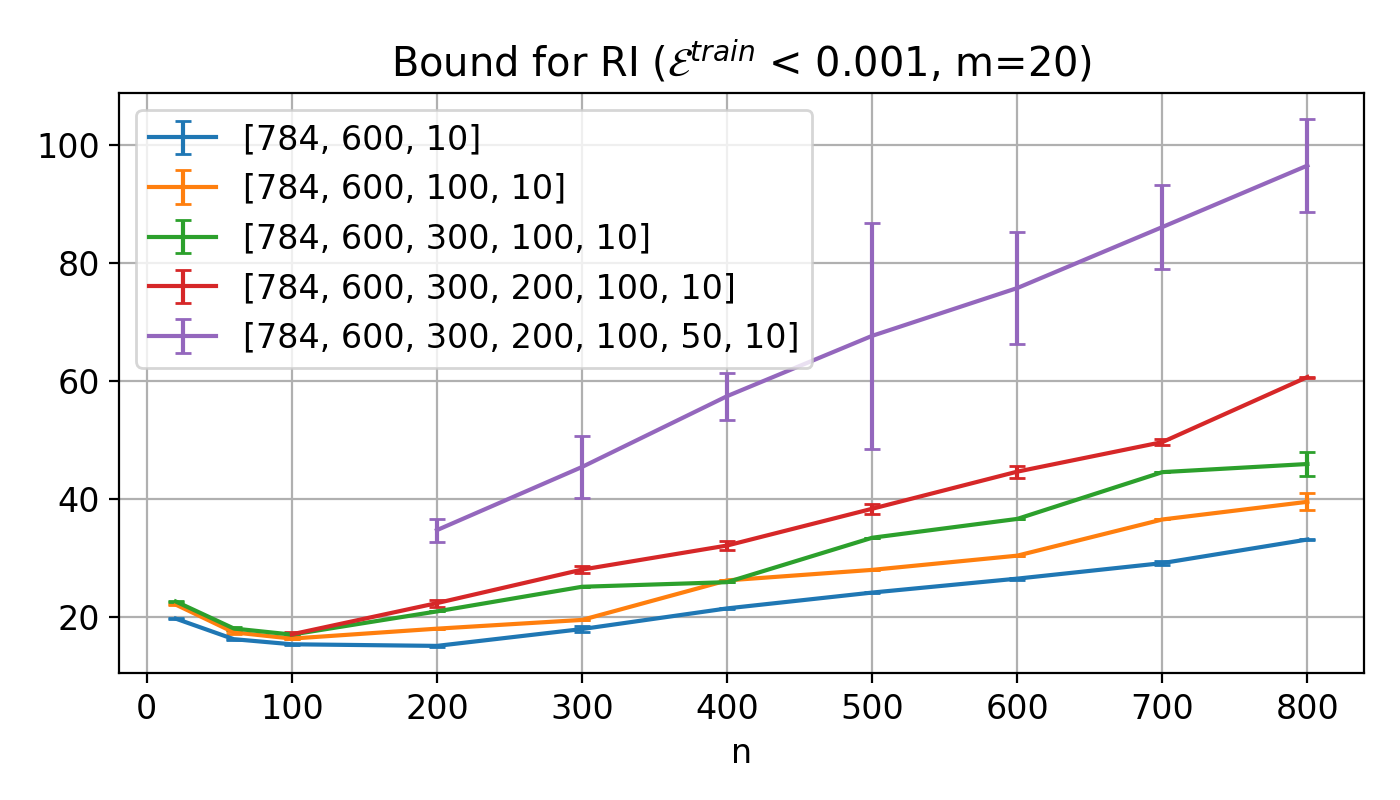}
    \caption{The (average) bound \eqref{exp-bound} computed for several different randomly
        initialized neural networks, with different architectures and for varying number
        $n$ of training samples. The missing data points for certain architectures
        indicate that the trained networks could not achieve $\mathcal{E}^{train} <
    0.001$.}
    \label{deep-nets}
\end{figure}

Finally, we recall that the product of norms $\Pi_{\ell =1}^{L+1} ||W_{\ell }||_{op}$ in
bound \eqref{exp-bound}
approximating the Lipschitz constant of $g_{\underline{\theta}}$ in 
bound \eqref{eq-generr-zero-1-0} can be very sensitive to depth. Indeed, one feature of Theorem
\ref{thm-C0-ReLU-1-0} is that the bound on test error for the  constructed network is
independent of architecture. In Figure \ref{deep-nets}, we see that although \eqref{exp-bound}
for randomly initialized networks does increase with depth, this change is less pronounced
the more strongly overparametrized the network is (i.e. small $n$).

\subsection{Interpretation of results}
We add the following remarks and observations on the experimental results:
\begin{itemize}
\item
The MNIST data set of handwritten digits contains a large number of pixels which are identical (white background) for all samples. This implies that the data set is contained in a linear subspace of $\R^{784}$ dimension smaller than 784. The rank of the data matrix $X_0$ increases with the number of training samples $n$ until $n\approx550$, and stabilizes around that value as $n$ further increases. This implies that the MNIST data set is contained in a subspace of dimension $\approx 550$. 
\item
As our numerical experiments show, the theoretical zero loss bound deteriorates once $n>200$. This indicates that the bulk of the MNIST data set is concentrated on a subspace of dimension $\approx 200$ or less, while the excess dimensions $\approx 350=550-200$ account for random fluctuations. As a consequence, we estimate the "effective dimension" of input space to be only $\approx 200$ or less, which means that the training sample set should not exceed $\leq 20$ per digit (for 10 digits) to ensure that we are in the strongly overparametrized regime.
\item
For the reasons explained above, we infer that the regime $n<200$ is strongly overparametrized with high probability, and our theoretical bound is most effective here. {\em We emphasize that this is precisely the regime in which the VC dimension based bound \eqref{eq-VCbound-1-0} fails to be effective; moreover, our bound does not require the training and test distributions to be identical.} As $n>200$ increases, the strongly overparametrized regime is abandoned, and the theoretical bound (which assumes strong overparametrization) deteriorates, for both shallow and deep networks. 
\item 
The network TFZL is obtained by initializing gradient descent with the theoretical zero loss minimizer from Theorem \ref{thm-C0-ReLU-1-0}. We perform this calculation to provide numerical evidence that the latter is stable (is a fixed point) under gradient descent.
\item
We note that while our theoretical bounds \eqref{eq-generr-zero-1-1} and especially \eqref{exp-bound} may not be sharp, the range of values of the relative error $\gamma$ displayed in Table \ref{table} indicates that, with an average error margin of 22\%, their performance  is quite strong in the strongly overparametrized regime, for the class of zero loss minima constructed in Theorem \ref{thm-C0-ReLU-1-0}. They in particular show that the generalization error is bounded and independent of the network architecture.  
\item
The trained network RI obtained from random initialization produces (approximate) loss minimizers $\uZ_*^{RI}$ that can differ significantly from the theoretical zero loss minimizer $\uZ_*^{ZL}$ corresponding to \eqref{eq-generr-zero-1-1} in Theorem \ref{thm-C0-ReLU-1-0}, as the set of zero loss minima is degenerate. As shown in Table \ref{table}, the theoretical bound using $\uZ_*^{RI}$ instead of $\uZ_*^{RI}$ performs poorly because the bound \eqref{exp-bound} on $\Lip(g_{\uZ_*^{RI}})$ is suboptimal, while it is optimal for $\Lip(g_{\uZ_*^{ZL}})$. Improvements on the theoretical bounds on $\Lip(g_{\uZ_*}^{RI})$ will be addressed in future work.
\end{itemize}

\appendix

\section{Proof of Proposition \ref{thm-generr-1-0}}

To begin with, we have
\eqn\label{eq-errdiff-1-0}
	\cE_{\uZ_*}^{train} - \cE_{\uZ_*}^{test}  
	&=&
	\frac{1}{n }\sum_{i=1}^{n}
	|f_{\uZ_*}(x_{i}^{(0)})-y_i|^2 -
	\frac1m\sum_{j=1}^{m}|f_{\uZ_*}(\widetilde x_{j}^{(0)})-\widetilde y_j|^2
	\nonumber\\
	&=&
	\frac{1}{n }\sum_{i=1}^{n}
	\Big(
	|f_{\uZ_*}(x_{i}^{(0)})-y_i|^2 -
	\frac1m\sum_{j=1}^{m}|f_{\uZ_*}(\widetilde x_{j}^{(0)})-\widetilde y_j|^2
	\Big) 
	\nonumber\\
	&=&
	\frac{1}{n m}\sum_{i=1}^{n}\sum_{j=1}^{m}
	\Big(
	|f_{\uZ_*}(x_{i}^{(0)})-y_i|^2 -
	|f_{\uZ_*}(\widetilde x_{j}^{(0)})-\widetilde y_j|^2
	\Big) 
\eeqn
where we used $1=\frac{1}{n }\sum_{i=1}^{n}1$ to pass to the second line, and $1=\frac{1}{m}\sum_{i=1}^{m}1$ to pass to the third line. Thus, \eqref{eq-cE-diff-1-0} follows.

Next, we note that 
\eqn 
	\lefteqn{
	\Big|\frac{1}{n m}\sum_{i=1}^{n}\sum_{j=1}^{m}
	\Big(
	|f_{\uZ_*}(x_{i}^{(0)})-y_i|^2 -
	|f_{\uZ_*}(\widetilde x_{j}^{(0)})-\widetilde y_j|^2
	\Big)\Big|
	}
	\nonumber\\
	&=&
	\Big|\frac{1}{n m}\sum_{i=1}^{n}\sum_{j=1}^{m}
	\Big(
	(f_{\uZ_*}(x_{i}^{(0)})-f_{\uZ_*}(\widetilde x_{j}^{(0)}))
	+(\widetilde y_j-y_i)\Big)
	\nonumber\\
	&&\hspace{2cm}
	\cdot \Big((f_{\uZ_*}(x_{i}^{(0)})
	+f_{\uZ_*}(\widetilde x_{j}^{(0)}))
	-(\widetilde y_j+y_i)	\Big)\Big|
	\nonumber\\
	&\leq&
	\frac{1}{n m}\sum_{i=1}^{n}\sum_{j=1}^{m}
	\Big(
	|f_{\uZ_*}(x_{i}^{(0)})-f_{\uZ_*}(\widetilde x_{j}^{(0)})|
	+|\widetilde y_j-y_i|\Big)
	\nonumber\\
	&&\hspace{2cm}
	\Big(|f_{\uZ_*}(x_{i}^{(0)})
	+f_{\uZ_*}(\widetilde x_{j}^{(0)})|
	+|\widetilde y_j+y_i|	\Big)
\eeqn
Using $\frac{1}{n m}\sum_{i=1}^{n}\sum_{j=1}^{m}1=1$, we obtain the upper bound
\eqn 
	&\leq&
	\sup_{(x,y)\in\cS^{train}}
	\sup_{(\widetilde x,\widetilde y)\in\cS^{test}}
	\Big(
	|f_{\uZ_*}(x)-f_{\uZ_*}(\widetilde x)|
	+|\widetilde y-y|\Big)
	\nonumber\\
	&&\hspace{1cm}
	\sup_{(x,y)\in\cS^{train}}
	\sup_{(\widetilde x,\widetilde y)\in\cS^{test}}\Big(|f_{\uZ_*}(x)|
	+|f_{\uZ_*}(\widetilde x)|
	+|\widetilde y|+|y|	\Big)
\eeqn 
By Lipschitz continuity of $f_{\uZ_*}$, there exists a constant $c_0:=\Lip(f_{\uZ_*})$ such that
\eqn
	|f_{\uZ_*}(x)-f_{\uZ_*}(\widetilde x)| \leq c_0  
	|x-\widetilde x|
\eeqn
and
\eqn 
	|f_{\uZ_*}(x)-f_{\uZ_*}(0)| \leq c_0 |x|
\eeqn 
so that
\eqn 
	|f_{\uZ_*}(x)| \leq |f_{\uZ_*}(0)| + c_0 |x| \,.
\eeqn 
Hence, the above is bounded by
\eqn\label{eq-diam-bd-1-0}
	&\leq&
	\sup_{(x,y)\in\cS^{train}}
	\sup_{(\widetilde x,\widetilde y)\in\cS^{test}}
	\Big(
	c_0|x-\widetilde x|
	+|\widetilde y-y|\Big)
	\nonumber\\
	&&\hspace{1cm}
	\sup_{(x,y)\in\cS^{train}}
	\sup_{(\widetilde x,\widetilde y)\in\cS^{test}}
	\Big(2|f_{\uZ_*}(0)| + c_0 (|x|+|\widetilde x|) 
		+|\widetilde y|+|y|	\Big)
	\nonumber\\
	&\leq&
	(1+c_0)\sup_{(x,y)\in\cS^{train}}
	\sup_{(\widetilde x,\widetilde y)\in\cS^{test}}
	\Big(
	|x-\widetilde x|
	+|\widetilde y-y|\Big)
	\nonumber\\
	&&\hspace{1cm}
	2(1+|f_{\uZ_*}(0)| +c_0)\sup_{(x,y)\in\cS^{train}}
	\sup_{(\widetilde x,\widetilde y)\in\cS^{test}}
	\Big(1+ |x|+|\widetilde x|
		+|\widetilde y|+|y|	\Big)
	\nonumber\\
	&\leq&
	2\sqrt2(1+|f_{\uZ_*}(0)| +c_0)^2(1+2R)
	\sup_{(x,y)\in\cS^{train}}
	\sup_{(\widetilde x,\widetilde y)\in\cS^{test}}
	\sqrt{|x-\widetilde x|^2
	+|\widetilde y-y|^2 }
		\nonumber\\
	&\leq&
	C_0(1+R) \; {\rm diam}(\cS^{train}\cup\cS^{test})  
\eeqn 
where $R$ is the radius of the smallest ball centered at the origin of $\R^{M_0}\times\R^Q$ that contains $\cS^{train}\cup\cS^{test}$. 

We infer from the above steps that the constant $C_0$ is bounded by
\eqn\label{eq-C0-bd-1-0} 
	C_0 \leq 8(1+|f_{\uZ_*}(0)| + c_0)^2
\eeqn 
where $c_0$ is the Lipschitz constant of $f_{\uZ_*}$.
Control of $C_0$ in \eqref{eq-C0-bd-1-0} is obtained from Lemma \ref{lm-Lip-1-0}, below.
\qed

\begin{lemma}\label{lm-Lip-1-0}
	The Lipschitz constant $c_0$ of $f_{\uZ_*}$ is bounded by
	\eqn 
		c_0\leq c_\sigma^L\prod_{\ell=1}^{L+1}\|W_\ell^*\|_{op}
	\eeqn 
	where $c_\sigma$ is the Lipschitz constant of $\sigma$, $\|\;\cdot\;\|_{op}$ is the matrix operator norm, and $(W_\ell^*)_{\ell}$ are the minimizing weights for the training loss.
	
	Moreover, recalling the bounds \eqref{eq-sigma-bd-1-0} on $\sigma$,
	\eqn 
		|f_{\uZ_*}(0)|
		\leq C_1(a_0,a_1,\{\|W_\ell^*\|_{op},|b_\ell^*|\}_{\ell=1}^{L+1})
	\eeqn
	is bounded by a constant depending on the operator norms of the trained weights, and the Euclidean norms of the trained biases.
\end{lemma}

\prf
The map \eqref{eq-layermap-1-0}	in the $\ell$-th layer is Lipschitz continuous,
\eqn 
	\lefteqn{
	|\sigma(W_\ell^* x_1^{(\ell-1)}+b_\ell^*)- \sigma(W_\ell^* x_2^{(\ell-1)}+b_\ell^*)|
	}
	\nonumber\\
	&\leq& c_\sigma |W_\ell^* x_1^{(\ell-1)}-W_\ell^* x_2^{(\ell-1)}|
	\nonumber\\
	&\leq& c_\sigma \|W_\ell^*\|_{op} |x_1^{(\ell-1)}-x_2^{(\ell-1)}|
\eeqn 
with Lipschitz constant 	$c_\sigma \|W_\ell^*\|_{op}$. We also recall the basic fact that if a pair of functions $f_1,f_2$ has Lipschitz constants $c_1,c_2$ and the composition $f_1\circ f_2$ is well defined, then $f_1\circ f_2$ has a Lipschitz constant bounded by $c_1 c_2$. By recursion in $\ell=1,\dots,L+1$, we arrive at the assertion.

Finally, recalling \eqref{eq-sigma-bd-1-0}, we obtain
\eqn  
	|x^{(\ell)}|
	&\leq&|\sigma(W_\ell^* x^{(\ell-1)}+b_\ell^*)|
	\nonumber\\
	&\leq& a_0+a_1|W_\ell^* x^{(\ell-1)}+b_\ell^*|
	\nonumber\\
	&\leq& a_0+a_1(\|W_\ell^*\|_{op}|x^{(\ell-1)}|+|b_\ell^*|)\,.
\eeqn 
Using $x^{(0)}=0$, recursion implies that (recalling that the output layer contains no activation function)
\eqn 
	|f_{\uZ_*}(0)|&=&|x^{(L+1)}|
	\nonumber\\
	&\leq&\|W_{L+1}^*\|_{op}|x^{(L)}|+|b_{L+1}^*|
	\nonumber\\
	&\leq&\|W_{L+1}^*\|_{op}(a_0+a_1(\|W_L^*\|_{op}|x^{(L-1)}|+|b_L^*|))
	+|b_{L+1}^*|
	\nonumber\\
	&\leq&\cdots\;\leq\;
	C_1(a_0,a_1,\{\|W_\ell^*\|_{op},|b_\ell^*|\}_{\ell=1}^{L+1})
\eeqn 
is bounded by a constant depending on the operator norms of the trained weights, and the norms of the trained biases.
\endprf

\begin{remark}
The bounds proven in Lemma \ref{lm-Lip-1-0} are consistent with the experimental observation that smaller weights increase generalization performance, through the decrease of the constants $c_0$ and $C_1$ in the current setting. In machine learning, training often aims to minimize both the loss and a term proportional to the $L^2$ norm of all parameters. This can be achieved through weight decay, where this term is added to the loss function. Additionally, gradient-based methods are known to introduce this term as an implicit regularizer in some cases, \cite{galsieguppog,var}.
\end{remark}

\begin{remark}
	Generalization bounds involving the spectral norms of weights and Lipschitz constants of the activation have also been obtained in the context of probabilistic estimates, see for instance \cite{barfostel,nagkol-1,neybhomcasre-1} and the references therein.
\end{remark}

\section{Proof of Theorem \ref{thm-main-zeroloss-0-1}}

Let $g_{\uZ_*}$ be the map from the first hidden layer to the output layer for the trained network; hence, we have that
\eqn 
	f_{\uZ_*}(x) = g_{\uZ_*}(W_1^* x + b_1^*) \,.
\eeqn 
It follows that
\eqn\label{eq-fdiff-1-0}
	|f_{\uZ_*}(\widetilde x)-f_{\uZ_*}(x)| &\leq&
	|g_{\uZ_*}(W_1^* \widetilde x + b_1^*)-g_{\uZ_*}(W_1^* x + b_1^*)|
	\nonumber\\
	&\leq&
	\Lip(g_{\uZ_*}) |W_1^* (\widetilde x-x) | \,.
\eeqn 
We now consider each term of 
\eqn\label{eq-errdiff-1-2}
	\cE_{\uZ_*}^{test} 
	&=&
	\frac{1}{m}\sum_{j=1}^{m} 
	|f_{\uZ_*}(\widetilde x_{j}^{(0)})-\widetilde y_j|^2  .
\eeqn
We find that for any arbitrary training data point $(x_{i}^{(0)},y_{i})\in \cS^{train}$, we have
\eqn\label{eq-testtrain-bd-1-0}
	\lefteqn{
	|f_{\uZ_*}(\widetilde x_{j}^{(0)})-\widetilde y_j|^2 
	}
	\nonumber\\
	&=&|f_{\uZ_*}(\widetilde x_{j}^{(0)})-
	\widetilde y_j- (f_{\uZ_*}( x_{i}^{(0)})-y_{i}) +\underbrace{(f_{\uZ_*}(
    x_{i}^{(0)})-y_{i})}_{=: r(x_i^{(0)}, y_i)}|^2
	\nonumber\\
	&\leq &
	\big(|f_{\uZ_*}(\widetilde x_{j}^{(0)})-f_{\uZ_*}( x_{i}^{(0)})|
	+|\widetilde y_j-y_{i}| + |r(x_i^{(0)}, y_i)|\big)^2
	\nonumber\\
	&\leq&
 	\Big(\Lip(g_{\uZ_*}) |W_1^* (\widetilde x_{j}^{(0)}
    - x_{i}^{(0)})| +|\widetilde y_j-y_{i}| + |r(x_i^{(0)}, y_i)|\Big)^2  
	\nonumber\\
	&\leq&
 	(1 + \delta) \Big(\Lip(g_{\uZ_*}) |W_1^* (\widetilde x_{j}^{(0)}
    - x_{i}^{(0)})| +|\widetilde y_j-y_{i}|\Big)^2  + \left(1 +
    \frac{1}{\delta}\right)|r(x_i^{(0)}, y_i)|^2
	\nonumber\\
	&\leq&
	(1 + \delta) C_{\uZ_*}^2\Big( |W_1^* (\widetilde x_{j}^{(0)}
	- x_{i}^{(0)})| + |\widetilde y_j-y_{i}| \Big)^2 
	\nonumber\\
    &&\hspace{12em}+ \left(1 + \frac{1}{\delta}\right)|r(x_i^{(0)}, y_i)|^2
\eeqn 
with
\eqn 
	C_{\uZ_*}:=
	\max\{1,\Lip(g_{\uZ_*}) \} \,.
\eeqn 
Here, we used \eqref{eq-fdiff-1-0} to pass to the third line, and the fourth line follows
from the Peter--Paul inequality (or Young's inequality).

For each $j$, we now define $(x_{i(j)}^{(0)},y_{i(j)})\in \cS^{train}$ as the training
data point minimizing the first term in \eqref{eq-testtrain-bd-1-0},
\eqn 
	|W_1^* (\widetilde x_{j}^{(0)}
	- x_{i(j)}^{(0)})|
	+|\widetilde y_j-y_{i(j)}| 
	=\min_{i=1,\dots,n}
	\Big( |W_1^* (\widetilde x_{j}^{(0)}
	- x_{i}^{(0)})|
	+|\widetilde y_j-y_{i}| \Big) \,.
\eeqn 
Therefore, we obtain
\eqn  
	\eqref{eq-errdiff-1-2} 
	&\leq&
    (1 + \delta) C_{\uZ_*}^2\frac{1}{m}\sum_{j=1}^{m} \Big( |W_1^* (\widetilde x_{j}^{(0)}
	- x_{i(j)}^{(0)})|
	+|\widetilde y_j-y_{i(j)}| \Big)^2 
	\nonumber\\
	&&\hspace{2cm}+ \left(1+ \frac{1}{\delta}\right)
    r(x_{i(j)}^{(0)}, y_{i(j)})^{2} 
	\nonumber\\
	&=&
	(1 + \delta)C_{\uZ_*}^2 \distCD^2(\cS^{test}_{W_1^* }\,|\,
	\cS^{train}_{W_1^* }) + \left(1+ \frac{1}{\delta}\right)
    \mathcal{E}_{\underline{\theta}_{*}}^{testNN}
\eeqn
where $\distCD(\cS^{test}_{W_1^* }\,|\,
	\cS^{train}_{W_1^* })$ is the Chamfer pseudodistance between the point clouds $\cS^{test}_{W_1^* }$ and 
	$\cS^{train}_{W_1^* }$, and
\begin{align}
    \mathcal{E}_{\underline{\theta}_{*}}^{testNN} := \frac{1}{m}\sum_{j=1}^{m} 
    r(x_{i(j)}^{(0)}, y_{i(j)})^{2} 
\end{align}
is the average training error for points in $\mathcal{S}^{train}$ which minimize the
Chamfer distance to the test points.\footnote{Here, $NN$ stands for nearest neighbor. Note
that training points might contribute more than once to
$\mathcal{E}_{\underline{\theta}_{*}}^{testNN}$.}

If zero training loss is achieved, then 
\eqn 
	f_{\uZ_*}(x) = g_{\uZ_*}(W_1^* x + b_1^*) = y \,
\eeqn 
for all $(x,y)\in\cS^{train}$. This is equivalent to 
\eqn 
	W_1^* x + b_1^* = g_{\uZ_*}^{-1}(y)
\eeqn 
where the right hand side denotes a (possibly non-unique) preimage of $y$ in the first layer.
This implies that for the data matrix $X_0=[x_1\cdots x_n]$,
\eqn 
	W_1^* X_0   = A := [g_{\uZ_*}^{-1}(y_1)-b_1^*\;\;\cdots\;\; g_{\uZ_*}^{-1}(y_n)-b_1^*] 
\eeqn 
and hence,
\eqn 
	W_1^*    = A X_0^+ \,.
\eeqn 
For the orthogonal projector $P_{X_0}=X_0 X_0^+$ onto the range of $X_0$, we therefore find that $W_1^*  P_{X_0}=A X_0^+ (X_0 X_0^+)=A X_0^+ = W_1^*$ acts projectively, and is supported on the range of $X_0$, which equals the span of the training data.

This proves the theorem.
\qed

\section{Proof of Theorem \ref{thm-C0-ReLU-1-0}}
 
We first construct zero training loss minimizers for the $L^2$ loss function, using a similar argument as first presented in \cite{chemoo-1}. Recalling the notations \eqref{eq-Bell-def-1-0}, zero loss is attained if and only if in the output layer,
\eqn\label{eq-WXout-1-0} 
	W_{L+1}X^{(L)} = Y - B_{L+1} \,
\eeqn 
where $Y$
is the matrix of reference output vectors $y_i\in\R^Q$.
We choose 
\eqn 
	b_{L+1}=-2\alpha(1,\dots,1)^T
	\;\;\;,\;\;
	\alpha =  \max_{i,j}|(Y_{ij})_-| \,,
\eeqn 
so that
\eqn 
	Y - B_{L+1} \;\;\;\in \R^{Q\times n}_+
\eeqn 
has non-negative matrix components. In particular, if all components of $Y$ are non-negative, $\alpha=0$.

In general, we may choose an arbitrary
\eqn\label{eq-WL1-gen-1-0} 
	W_{L+1} \in \R^{Q\times M_L}_+
	\;\;\;,\;\;
	\rank(W_{L+1}) = Q
\eeqn 
of full rank, so that $W_{L+1}W_{L+1}^T\in\R^{Q\times Q}$ is invertible, with non-negative matrix components. 
Then, denoting the generalized inverse of $W_{L+1}$ by
\eqn
	W_{L+1}^+ := W_{L+1}^T(W_{L+1}W_{L+1}^T)^{-1} 
	\;\;\;\in\;\R^{M_{L}\times Q}_+
\eeqn
we obtain that
\eqn\label{eq-XL-sol-1-0} 
	X^{(L)} = W_{L+1}^+(Y - B_{L+1})
	\;\;\;\in \R^{M_L\times n}_+
\eeqn
satisfies \eqref{eq-WXout-1-0}, and is guaranteed to have non-negative matrix components.

In the case at hand, it is convenient to make the explicit choice
\eqn 
	W_{L+1} = [\1_{Q\times Q} \;\;0_{Q\times (M_{L}-Q)}]
\eeqn 
with generalized inverse
\eqn\label{eq-WL1-geninv-1-0}
	W_{L+1}^+ = 
	\left[
	\begin{array}{c}
	\1_{Q\times Q}\\
	0_{(M_{L}-Q) \times Q}
	\end{array}
	\right] \,
\eeqn 
where $0_{p\times q}\in\R^{p\times q}$ consists of zeros.
We then find for \eqref{eq-XL-sol-1-0} 
\eqn 
	X^{(L)} = W_{L+1}^+(Y-B_{L+1})
	=
	\left[
	\begin{array}{c}
	Y-B_{L+1} \\
	0_{(M_{L}-Q)\times n}
	\end{array}
	\right] 
	\;\;\;\in\R^{M_{L}\times n}_+\,.
\eeqn 
In particular, $W_{L+1}$ and $W_{L+1}^+$ have operator norms
\eqn\label{eq-WL1-bds-1} 
	\|W_{L+1}\|_{op} \;=\;
	\|W_{L+1}^+\|_{op} \;=\; 1\,. 
\eeqn 
The subsequent steps are the same for this specific choice of $W_{L+1}$, or for an arbitrary choice of the form \eqref{eq-WL1-gen-1-0}.

In the next higher, $L$-th layer, we have
\eqn 
	X^{(L)}= \sigma(W_L X^{(L-1)} +B_L)
	\;\;\;\in \R^{M_L\times n}_+
\eeqn 
and choose
\eqn 
	W_L&=&[\1_{M_L\times M_L} \;\;0_{(M_{L-1}-M_L)\times M_L}]
	\;\;\;\in\R^{M_L\times M_{L-1}}_+
	\nonumber\\
	b_L&=&0\,,
\eeqn 
and we let
\eqn 
	X^{(L-1)}= 
	\left[
	\begin{array}{c}
	X^{(L)} \\
	0_{(M_{L-1}-M_L)\times n}
	\end{array}
	\right] 
	\;\;\;\in\R^{M_{L-1}\times n}_+\,.
\eeqn 
By recursion, we choose
\eqn 
	W_\ell&=&[\1_{M_\ell\times M_\ell} 
	\;\;0_{(M_{\ell-1}-M_\ell)\times M_L}]
	\;\;\;\in\R^{M_\ell\times M_{\ell-1}}_+
	\nonumber\\
	b_\ell&=&0\,,
\eeqn 
for $\ell=2,\dots,L$, so that
\eqn 
	X^{(\ell)}= 
	\left[
	\begin{array}{c}
	X^{(L)} \\
	0_{(M_{\ell}-M_L)\times n}
	\end{array}
	\right] 
	\;\;\;\in\R^{M_{\ell}\times n}_+\,.
\eeqn 
%
Hence, in the layer $\ell=1$, we arrive at
\eqn\label{eq-X1-sol-1-0} 
	X^{(1)}= 
	\left[
	\begin{array}{c}
	X^{(L)} \\
	0_{(M_{1}-M_L)\times n}
	\end{array}
	\right] 
	\;\;\;\in\R^{M_{1}\times n}_+\,.
\eeqn 
and
\eqn\label{eq-X1eq-1-0}
	X^{(1)} = \sigma(W_1 X^{(0)}+B_1) \,.
\eeqn 
Here, we note that since $X_0\equiv X^{(0)}\in\R^{M_0\times n}$ has full rank $n$, it follows that the matrix $X_0^TX_0\in\R^{n\times n}$ is invertible, and we solve \eqref{eq-X1eq-1-0} with
\eqn 
	W_1=X^{(1)}X_0^+
	\;\;\;,\;\;
	b_1=0 \,,
\eeqn 
where 
\eqn 
	X_0^+=(X_0^TX_0)^{-1} X_0^T \,
\eeqn 
denotes the generalized inverse of $X_0$.
Recalling \eqref{eq-WL1-geninv-1-0}, we have
\eqn
	W_1 &=& 
	\left[
	\begin{array}{c}
		W_{L+1}^+ (Y - B_{L+1})
	X_0^+
	\\
	0_{(M_{1}-M_L)\times M_0}
	\end{array}
	\right]
	\nonumber\\
	&=&
	\left[
	\begin{array}{c}
	(Y - B_{L+1})
	X_0^+
	\\
	0_{(M_{1}-Q)\times M_0}
	\end{array}
	\right]
	\;\;\;\in\;\R^{M_1\times M_0}
\eeqn
by collecting rows of zeros. While $W_1$ has entries of indefinite sign,  the product
\eqn 
	W_1 X_0 = \left[
	\begin{array}{c}
	Y - B_{L+1}
	\\
	0_{(M_{1}-Q)\times n}
	\end{array}
	\right]
	\;\;\;\in\;\R^{M_1\times n}_+
\eeqn 
has non-negative components. Therefore, $\sigma$ acts on the latter as the identity, and we indeed verify that \eqref{eq-X1eq-1-0} is satisfied with \eqref{eq-X1-sol-1-0} and $b_1=0$.

Therefore, we arrive at the explicit zero loss minimizer $\uZ_*\in\R^K$ consisting of
\eqn 
	W_1^* &=& 
	\left[
	\begin{array}{c}
	(Y - B_{L+1}^*)
	X_0^+
	\\
	0_{(M_{1}-Q)\times M_0}
	\end{array}
	\right]
	\nonumber\\
	W_\ell^*&=&[\1_{M_\ell\times M_\ell} \;\;0_{(M_{\ell-1}-M_\ell)\times M_L}]\;\;\;,\;\;\ell=2,\dots,L
	\nonumber\\
	W_{L+1}^* &=& [\1_{Q\times Q} \;\;0_{Q\times (M_{L}-Q)}]
	\nonumber\\
	b_\ell^*&=&0\;\;\;,\;\;\ell=1,\dots,L
	\nonumber\\
	b_{L+1}^*&=&-\alpha(1,\dots,1)^T
	\;\;\;,\;\;
	\alpha =  \max_{i,j}|(Y_{ij})_-| \,.
\eeqn  
Then, similarly as in Lemma \ref{lm-Lip-1-0}, we find that the Lipschitz constant of $g_{\uZ_*}$ is bounded by
\eqn 
		\Lip(g_{\uZ_*})\leq
		c_\sigma^L\prod_{\ell=2}^{L+1}\|W_\ell^*\|_{op}
		= 1 \,,
\eeqn
where we used \eqref{eq-WL1-bds-1}, and the fact that the Lipschitz constant for ReLU is $c_\sigma=1$. 
Therefore, the constant in \eqref{eq-CuZ-Lip-1-0} satisfies
\eqn 
	C_{\uZ_*} &=& \max\{1,\Lip(g_{\uZ_*}) \}
	\nonumber\\
	&=&1 \,. 
\eeqn 
%
%
The generalization bound then follows from \eqref{eq-generr-zero-1-0}.
This proves the claim.
\qed

$\;$
 
\noindent
{\bf Acknowledgments:} 
T.C. thanks Adam Klivans and Vardan Papyan for inspiring discussions, and gratefully acknowledges support by the NSF through the grant DMS-2009800, and the RTG Grant DMS-1840314. P.M.E. was supported by the NSF through T.C. with the grant DMS-2009800.
\\


\newcommand{\etalchar}[1]{$^{#1}$}

\end{document}